\documentclass{article}
\usepackage[T1]{fontenc}
\usepackage[utf8]{inputenc}
\usepackage{graphicx}
\usepackage{mathtools}
\usepackage{amsmath}
\usepackage{amssymb}
\usepackage{mathrsfs}
\usepackage{mathtools}
\usepackage[makeroom]{cancel}
\graphicspath{ {Images/} }
\usepackage{xcolor}

\usepackage{hyperref}
\usepackage{rotating}
\usepackage{verbatim}
\usepackage{caption}
\usepackage{subcaption}
\usepackage{fancyvrb}
\usepackage{enumerate}
\usepackage{relsize}
\usepackage{float}

\usepackage[square]{natbib}
\setcitestyle{numbers}
\usepackage{pdflscape}

\usepackage{tikz}
\usetikzlibrary{fit,positioning}

\usepackage{hyperref}
\usepackage[margin=1.5in]{geometry}
\hypersetup{colorlinks,citecolor=blue,urlcolor=blue,linkcolor=blue}

\author{Jonathan Johannemann \\ \texttt{jonjoh@stanford.edu}
\and Robert Tibshirani \\ \texttt{tibs@stanford.edu}}
\date{Stanford University}
\title{Spectral Overlap and a Comparison of Parameter-Free, Dimensionality Reduction Quality Metrics}

\begin{document}

\maketitle

\begin{abstract}
    Nonlinear dimensionality reduction methods are a popular tool for data scientists and researchers to visualize complex, high dimensional data. However, while these methods continue to improve and grow in number, it is often difficult to evaluate the quality of a visualization due to a variety of factors such as lack of information about the intrinsic dimension of the data and additional tuning required for many evaluation metrics. In this paper, we seek to provide a systematic comparison of dimensionality reduction quality metrics using datasets where we know the ground truth manifold. We utilize each metric for hyperparameter optimization in popular dimensionality reduction methods used for visualization and provide quantitative metrics to objectively compare visualizations to their original manifold. In our results, we find a few methods that appear to consistently do well and propose the best performer as a benchmark for evaluating dimensionality reduction based visualizations.
\end{abstract}

\section{Introduction}

In a variety of modern applications, researchers find dimensionality reduction as a beneficial way of visualizing high dimensional data. Some examples of successful applications include HIV analysis \cite{betechuoh2006autoencoder} and analyzing gene data \cite{ekins2006insights} while some researchers might use dimensionality reduction to showcase algorithmic output as opposed to the raw data \cite{ying2018graph}, \cite{mikolov2013distributed}. But, while dimensionality reduction for visualization is often regarded as a helpful way to explore data, a quantitative measure of the low dimensional output's similarity to its high dimensional input is infrequently mentioned.

The task of dimensionality reduction for a data set $X^{n \times d}$ can be specified as follows. The researcher begins with a set of $n$ observations with $d$ dimensions which can be represented as $X = \{x_1,...x_n\}$. Next, a target dimension $p$ is chosen which could be any $p$ such that $p < d$ to reduce the number of unfavorable properties that come with high dimensional spaces \cite{jimenez1998supervised} or $p$ can be set to 2 or 3 dimensions as a means to visualize the high dimensional data. The dimensionality reduction algorithm is then applied to map $X^{n \times d}$ to a low dimensional set $Y^{n \times p} = \{y_i,...y_n\}$ while seeking to  maintain as much of the original structure from $X$ in $Y$. This problem is difficult in real applications due to a variety of reasons such as lack of information about the effective dimension or the geometry of the manifold on which the data lives \cite{van2009dimensionality}.

One of the challenges for researchers using tunable nonlinear dimensionality reduction algorithms is determining which performance metric in the literature is best suited for their application. The first difficulty in the current literature is the prevalence of quality metrics that require additional tuning from the user. Without a proper prior, this task can be difficult because different choices of metric parameter values result in different optimal choices for dimensionality reduction hyperparameters. Then, improper tuning of a quality metric can lead to misleading visualizations. The second is the lack of lower dimensional exploration of quality metric performance to ensure that the a given metric is useful for hyperparameter optimization. In the experiments section, we seek to compare these methods for more trivial cases in low dimensions to explore each method's ability to act as a performance metric when we can visualize the ground truth for the low dimensional local structure.

%which are the prevalence of tunable performance metrics and the lack of  Often times, quality metrics will require that the user provide additional hyperparameters for the metrics which specify the user's tolerance for inconsistencies in characteristics that were present in the high dimensional input but not low dimensional output. This requires the user to define locality which may be difficult without a prior that directly addresses these traits since different hyperparameters for the metrics will yield different optimal hyperparameters for the dimensionality reduction algorithms as well. Additionally, while there are performance metrics 

As the primary target of dimensionality reduction evaluation literature, nonlinear dimensionality reduction methods continue to grow in number and have succeeded in a variety of applications. The following methods are just a few of the algorithms that we use in our experiments to assess the quality metrics. An early method is Sammon mapping \cite{sammon1969nonlinear} where squared differences in high and low dimensional pairwise distances are scaled by the Euclidean distance in the original space $X$. Another very popular method, t-SNE \cite{maaten2008visualizing}, uses pairwise Euclidean distances to generate conditional probabilities in the high and low dimensional space and then seeks to minimize the KL divergence between a high dimensional Gaussian and low dimensional Student t-distribution with 1 degree of freedom. Also, \cite{chen2009local} propose their method ``local multidimensional scaling'' which utilizes a combination of MDS and concepts from ``force-direct'' graph drawing. Our emphasis is especially on methods such as local MDS and t-SNE which both have hyperparameters that can dramatically change the visualization and mislead a user into believing that there is additional underlying structure or that the classes are very easily separable. 

The contributions of the paper are as follows. We provide a joint comparison of recent ``parameter-free'' quality metrics and promising nonlinear dimensionality reduction algorithms for visualization. Based on our findings, we recommend $Q_{NX}$ \cite{liang2017new} and the Spectral Overlap quality metric developed in this paper.

The remainder of the paper is outlined as follows. In Section 2, we introduce existing challenges and past approaches. In Section 3, we discuss the quality metric literature and some of the pitfalls. Finally, in Section 4, we run experiments to evaluate popular and promising nonlinear dimensionality reduction algorithms for visualization using quality metrics that do not require the researcher to specify parameters such as the number of relevant $k$ nearest neighbors.

\section{Algorithms}

In some of the more basic dimensionality reduction algorithms for visualization such as multidimensional scaling and PCA, there are no additional tunable parameters to influence the outcome of the output \textit{Y} and the means by which one can interpret the data is straightforward. However, these linear techniques are limited and suffer from various pitfalls such as overly focusing on large pairwise distances and therefore compromise small pairwise distances which make up the local structure of the data \cite{van2009dimensionality}. In addition, some of the most successful applications of dimensionality reduction in areas such as computer vision, natural language processing, and audio signal processing have made tremendous strides through the use of highly nonlinear models. Both of these occurrences suggest that there is potentially more to gain by using nonlinear methods which focus on local structure in the input data.

But, since these methods are nonlinear and many do not develop an explicit mapping, additional concerns arise including the stability of visualizations and how well the newly generated output dimension represents the original high dimensional space. Interestingly, \cite{bunte2012general} tackle this by posing dimensionality reduction as an optimization problem and learn mapping functions which could potentially act as one way to visualize the stability of the dimensionality reduction process. 

%I probably want to examine this paper a little more closely to see if there's the problem of accidentally missing a subgroup when sampling from a large dataset to generate these mappings

Another area of difficulty for these nonlinear dimensionality reduction methods is that they are often favored for high dimensional tasks where there is no known equivalent to a ``ground truth''. In regression tasks, we have a ``ground truth'' response $y \in \mathbb{R}$ and in classification tasks, we have $y \in \mathbb{Z}^+$. Since the field of dimensionality reduction does not have a universal and similarly objective method for comparison, there is a reliance on the researcher to provide the remaining assumptions to correctly map some high dimensional X to low dimensional Y. We can see that this is the case in \cite{maaten2008visualizing} and \cite{mcinnes2018umap} where much of the analysis is qualitative and the evaluation is primarily based on an algorithm's ability to separate like-classes into separate clusters or groups. However, no numerical measurement is used to bolster the qualitative claims and provide formal comparison with respect to the original input dataset. This can be a bit concerning if it turns out that a visualization was spurious or has little to no relation to the original high dimensional data. For cases where the data may be used downstream in tasks like regression or classification, it might be reasonable to have some measure of similarity to the high dimensional space in order to better gauge if observed group separability or low dimensional structure is robust to new data.

%even in some cases, the estimated ''intrinsic dimension'' is greater than 2 or 3 dimensions and so that further makes the idea of finding the best or ''correct'' compression of that space subjective to some???? 

A demonstration conducted by \cite{wattenberg2016how} expresses the same concern by displaying how changing just one of t-SNE's tuning parameters can result in a variety of possible misleading visualizations. The authors discuss how, when one is using t-SNE, cluster sizes, distances between clusters, and interesting geometry in the visualization might be random or inconsistent with the original input data. Furthermore, they show how smaller values of t-SNE's ``perplexity'' hyperparameter, which is a smooth equivalent to a kNN graph, can result in different visualizations after each run of the algorithm. 

Ultimately, as we see in Section 3, the approach we propose is similar to the intuition behind the $K_{max}$ heuristic \cite{lee2010scale} which is to maintain as much local structure as possible but with a secondary consideration for global structure.

\section{Metrics}

The beginning of dimensionality reduction quality metrics dates back to the 1960's with methods such as Kruskal's Stress Measure and Sammon Stress. Since then, numerous methods have been developed to evaluate the effectiveness of an algorithm's ability to replicate high dimensional structure in a lower dimensional space \cite{gracia2014methodology} with an added emphasis on local structure. Many of these methods, such as trustworthiness and continuity \cite{venna2006local}, local continuity meta-criterion \cite{chen2009local}, and mean relative rank errors \cite{lee2007nonlinear} depend on an additional tunable parameter $k$ which requires the user to specify performance by an algorithm's ability to maintain the same $k$ nearest neighbors in some fashion. While a subjective $k$ might not mean much for a small number of data points, it can become unclear as $n \to \infty$ if a carefully chosen value for $k$ is reasonable.

Cognizant of this concern, many authors such as \cite{lee2008rank}, \cite{lee2010scale}, and \cite{liang2017new}, have tried evaluating projections qualitatively by plotting the curve created by evaluating tunable metrics for multiple values of $k$. Some complications with this approach can be seen in \cite{lee2010scale} if one reviews figure 5 which plots dimensionality reduction performance on the swiss roll dataset and figure 6 which plots dimensionality reduction performance on 1000 images drawn from the MNIST dataset. We see an immediate local maximum in the performance curve followed by a decline in values before increasing once again for large values of the parameter $k$. CCA using geodesic distances appears to be the best in figure 5 but t-SNE using Euclidean distance is labeled the best performer based on their rule of judging performance by the best ``local'' score. We see what appears to be the first few neighbors are kept very well but that same level of performance is not achieved until $k$ is much greater than the first location of the local maximum. This phenomenon occurs when there is an overemphasis on local neighborhoods such as in cases in \cite{wattenberg2016how}. In the examples with small perplexity, t-SNE forces immediate neighbors to be close but results in tiny clusters that have very little resemblance with the original dataset.

In addition, \cite{lee2009quality} propose the co-ranking matrix which is a more comprehensive means of evaluating the dimensionality reduction process. The co-ranking matrix is an \textit{(N-1)-by-(N-1)} matrix that is a joint histogram of the ranks with observations above the diagonal called ``extrusions'' and below the diagonal ``intrusions''. If used as an evaluation tool, one can identify hard intrusions and extrusions based on a choice of $k$ which determines the nearest neighbors. Lee and Verleysen also show how T\&C, MRRE, and LCMC are similar to penalizing different portions of the co-ranking matrix and go on to offer the co-ranking matrix as a framework for future development of dimensionality reduction quality metrics. Some of the insights gained through visualizing dimensionality reduction performance through this lens include allowing the user to identify less harmful, small intrusions/extrusions which may come with noise flattening and large amplitude extrusions which characterizes a tearing in the manifold.

While tunable metrics do provide substantial information about the relationship between high and low dimensional space, quality metrics without tunable parameters require fewer assumptions on the part of the user and therefore one can guarantee more consistency in the quality of the output. \cite{bezdek1995index} propose using a metric that measures preservation of distance orderings and have offered the Spearman's Rho as one of the early measures to determining the preservation of topology. Some newer methods have come out such as ``entropy'' and ``mutual information'' which were proposed by \cite{babaee2013assessment} and treat the dimensionality reduction process as a communication channel model which transfers data points from high to low dimensional space.

\section{Spectral Overlap}

Building on this understanding of nonlinear dimensionality reduction quality metrics, we propose the following method which we call ``Spectral Overlap''. The intuition here is that we want to penalize any lack of overlap in every KNN graph for $k=1,...,n-1$. This provides equal weight across all neighbors. 

We specify ``Spectral Overlap'' as follows: Let $\tilde{K}^i$ be the KNN graph in input space $X$ with $k$ parameter $i$ and $K^i$ be the KNN graph in output space $Y$ with $k$ parameter $i$. We begin by calculating the overlap penalty:
\begin{align}
    \text{Overlap} = \sum_{j=1}^{(n-1)} \sum_{k < j} \tilde{K}^i_{jk} * K^i_{jk}
\end{align}
The overlap measure above captures the mismatch in high and low dimensional KNN graphs. From here, we scale by a normalizing constant to characterize the decay in performance as one increases the number of data points. This yields the quality metric:
\begin{align}
    \text{Spectral Overlap} = \frac{\text{Overlap}}{\tfrac{n^2 (n-1)}{2}} = \frac{2}{n^2 (n-1)} \sum_{j=1}^{(n-1)} \sum_{k < j} \tilde{K}^i_{jk} * K^i_{jk}
\end{align}
If there is a tear in the manifold, more nearest-neighbor relationships will not be upheld and the penalty will be larger. Additionally, this also penalizes cases where one group can potentially occlude another due to the limits of the dimensionality reduction algorithm. We find that for linear dimensionality reduction algorithms, this can happen often such as in the case of the clusters data set we explore in Section \ref{sec:experiments}.

%This is different from $Q_{NX}$ \cite{liang2017new} which solely rewards lower dimensional projections on whether or not the high and low dimensional order rankings match and provides no gauge of the degree of mismatch. As we will see in the results section, methods that incorporate all pairwise rank order relationships appear to be more robust.

%Another similar method is nonmetric MDS which seeks to maintain the rank order of the pairwise distances. This is different because it seeks to maintain the unconditional ranking of the distances whereas spectral overlap seeks to maintain the ranking conditional on a given point $x_i$. Nonmetric MDS does not directly maintain KNN graphs and therefore we see some of the similar errors in our experiments, such as in \ref{fig:hdclus}, that occur with linear methods such as occlusion of groups and less focus on local structure.

Ultimately, due to the limit of metric spaces, spectral overlap's objective is to measure the preservation of nearest neighbor relationships and cannot properly address a lack of transitivity in cases such as word embeddings \cite{van2012visualizing}.  Thus, we alternatively recommend representing high dimensional data in multiple maps if the transitivity of data points is not guaranteed.

\section{Experiments}\label{sec:experiments}

To evaluate some of these more objective measures of local structure performance, we evaluate four methods on six data sets using 3 publicly available CRAN packages and a package we developed for local multidimensional scaling. We apply t-SNE from the \texttt{R} package {\fontfamily{qcr}\selectfont Rtsne} \cite{maaten2008visualizing}, UMAP from \texttt{umap} \cite{mcinnes2018umap}, Sammon Mapping from \texttt{stats} \cite{sammon1969nonlinear}, and local MDS from our package \texttt{lmds} \cite{chen2009local}. Next, while datasets such as the synthetic Swiss roll dataset \cite{tenenbaum2000global} and popular MNIST dataset \cite{lecun1998gradient} are common datasets for comparing nonlinear dimensionality reduction visualizations, we instead propose a series of datasets where we know confounding attributes such as the intrinsic dimension and local structure and utilize a Bayes error metric that complements each specific data generating process to evaluate the performance of the more general quality metrics that have been proposed. 

\subsection{Datasets}
The six datasets are: (1) Two Lines, (2) Trefoil Knot, (3) Three Gaussians, (4) Noisy Circles, (5) Curved X's, and (6) High Dimensional Clusters. Datasets (1) and (2) are directly from \cite{wattenberg2016how} and (3) is inspired by the three cluster dataset except we increase the variance for one cluster. Dataset (4) comes from \cite{sklearn2018comparing} and (5) is inspired by the parabola example in \cite{hastie2009elements}. Finally, we add (6) which is a set of separable clusters on the corners of a 4-dimensional hyper-cube.

\begin{sidewaysfigure}
  \includegraphics[width=\linewidth]{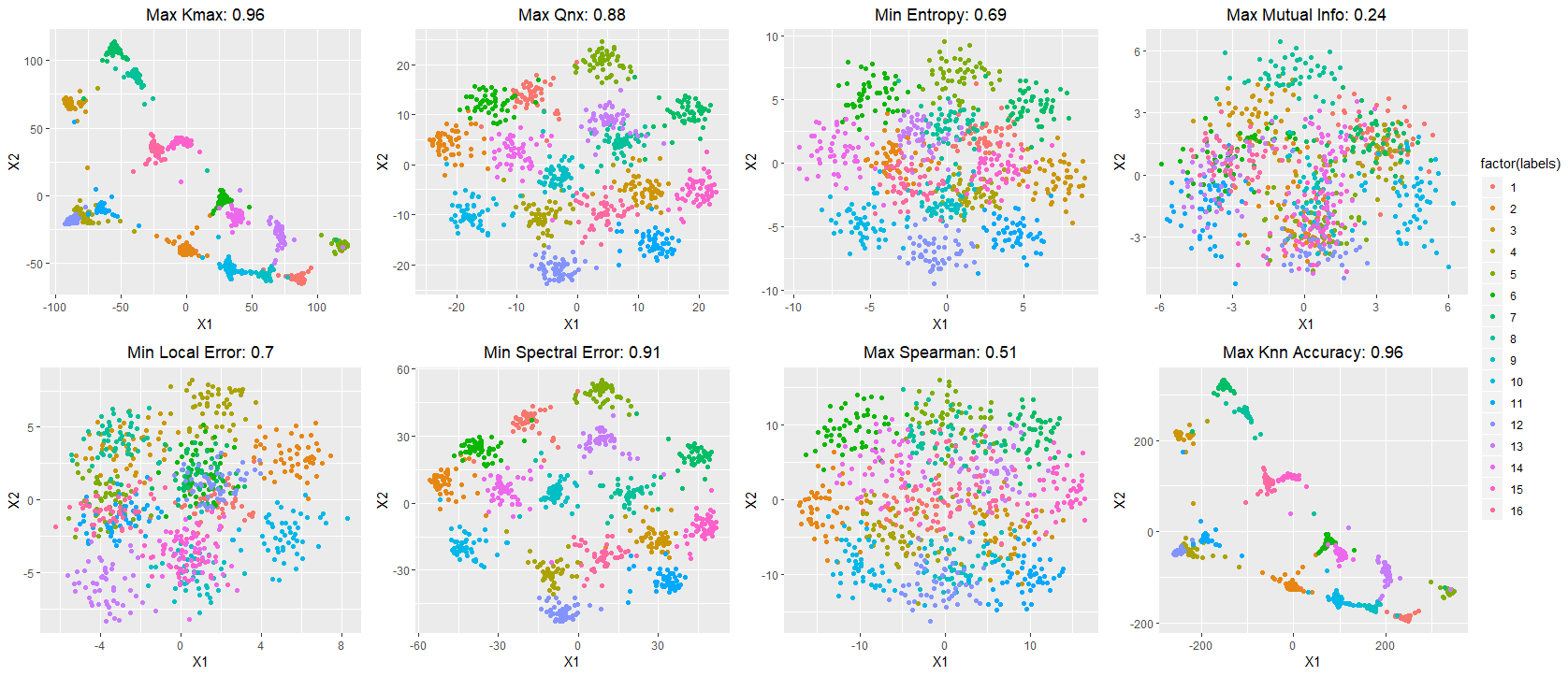}
  \caption{LMDS High Dimensional Clusters Result. Titles correspond to quality metrics optimized and the resulting $1-NN$ performance.}
  \label{fig:hdclus}
\end{sidewaysfigure}

\subsection{Setup}
Since we have the data generating processes for each method, we begin by generating 250 data points for each 2D dataset and 800 data points for the high dimensional cluster data set from the same random seed. Next, for each nonlinear dimensionality reduction algorithm, we pre-specify a range of values for each algorithm's set of hyperparameters and create a grid of all combinations for a grid search for each quality metric. The list of hyperparameters for each method is provided in the \texttt{lmds} github \href{https://github.com/JonathanJohann/lmds}{here}.

For each algorithm, we then set 5 different randomized seeds and evaluate each algorithm with a given group of hyperparameters from its corresponding grid using each of the different quality metrics. We then select the best scoring hyperparameter configurations for each algorithm and quality metric and re-run those algorithms with the same seed. These resulting $Y^{n \times p}$ visualizations are then representative of good performance for each nonlinear dimensionality reduction algorithm's parameters with respect to each quality metric.

Finally, we evaluate each algorithm-quality metric pair for each data set using Bayes error metrics that are appropriate for the data generating processes. For the 2-dimensional data sets, since we are projecting back to the same dimension, we are able to use the Procrustes Distance between the input data $X^{n \times d}$ and output visualization $Y^{n \times p}$ where $p = d$. The Procrustes Distance \cite{hastie2009elements} compares two data sets of dimension $X^{n \times d}$ and provides the L2 distance after scaling, rotating, and shifting one of the $X^{n \times d}$ data sets. This provides a means for measuring how well an algorithm can exactly replicate the original manifold if we are mapping to the intrinsic dimension of the data and know its topology exactly. Next, for the high dimensional clusters data sets, we use the accuracy of the first nearest neighbor in $Y^{n \times p}$ in classifying the class we assigned in high dimensional space $X^{n \times d}$. Once again, this is only relevant given that we have specifically devised a case where the groups are completely separable with no overlap. In both cases, the Bayes error metrics are hypothetical and complement the specific data generating process. However, by identifying these base cases, we can compare more general quality metrics such as those in  Appendix \ref{app:metrics} to get a better sense of robustness of said metrics prior to using them in more complex cases.

\subsection{Results \& Discussion}
Results from our simulations are shown in \ref{table:metric_summary.tex}. Since scale varies across datasets, we rank the performance of each metric on each algorithm-dataset pair from 1 to $m$ for the $m$ metrics including the ``Bayes error'' metric. Average performance is broken down by dimension first and then the cumulative performance for 2-dimensional and higher dimensional performance is averaged to generate a final score. 

Based on our simulations, we see that $Q_{NX}$ and Spectral Overlap tend to do the best in reducing the defined Bayes error metrics for each dataset and algorithm. This is then followed by Local Error and then Entropy. We see that $K_{max}$ which is based on a heuristic for tunable quality metrics and Mutual Information perform the worst. 

For the 2-dimensional data sets, we see that on average, across these metrics, Sammon mapping does the best followed by local MDS, then t-SNE, and finally UMAP. It appeared that local MDS, UMAP, and t-SNE had difficulty with the Curved X's dataset the most with higher Procrustes distances on average than the other datasets. The next worse dataset was the 3 Gaussians dataset. For a more visual exploration of the results, we provide visualizations of the original datasets as well as the best performers for each algorithm-quality metric pair in the supplementary material. 

However, on the multi-dimensional cluster data set, we see that Sammon mapping performs the worst and that more nonlinear methods tend to outperform across almost all quality metrics. Across most metrics, t-SNE appears to have the most robust performance with UMAP following closely. We see that $Q_{nx}$ performs the best followed by Local Error and Spectral Overlap. Perhaps one of the most interesting aspects of these results can be seen in the visualization in Figure \ref{fig:hdclus}. While $K_{max}$ has good numerical performance based on our choice of Bayes error, we can see that this Bayes error is prone to tearing of the manifold. $Q_{nx}$ and Spectral Overlap sought to maintain the relationship such that clusters were relatively equidistant from each other. While these did not perform as well, we can see that they did a better job of preserving what one could conceptualize as the original manifold.
\begin{table}[htbp]
	\scriptsize
	\centering
	\begin{tabular}{|c|c|c|c|}
\hline
Metric & 2D Avg. Rank & High Dim Avg. Rank & Overall Avg.\\
\hline
$K_{max}$  &5.88 & 4.50&5.19\\
$Q_{NX}$ & 4.28& \textbf{3.13} & \textbf{3.70}\\
Entropy & 4.38&5.13 &4.75\\
Mutual Info & 6.63 & 6.75&6.69\\
Local Error & 4.60 & 4.25&4.43\\
Spectral Error & 4.23 & 4.38&4.30\\
Spearman & \textbf{3.83} & 6.00&4.91\\
Procrustes/KNN & 2.20 & 1.88&2.04\\
\hline
\end{tabular}
	\caption{Average Ranked Performance}
	\label{table:metric_summary.tex}
\end{table}

Based on the outcomes, we see that $K_{max}$ is predisposed to certain edge cases that can mislead a user to think he or she has found a high quality visualization. As stated in Section 3, the problem appears in the case where an algorithm puts too much weight on maintaining a small number of $k$ nearest neighbors but subsequently compromises the remaining neighbor rankings. We see that, on average, $Q_{nx}$ and Spectral Overlap perform well in both the low and high dimensional data sets which leads us to believe for now that these are good proxies for quality of output dimension $Y$.

Ultimately, this demonstration shows how tuning nonlinear dimensionality reduction methods using various meta-criterion perform in the most ideal case when we know (1) the intrinsic dimension, (2) the true local structure or any topological structure, and (3) the true global structure. 

\section{Conclusion}

We present a simple exploration of some of the existing quality metrics that researchers can use without a prior with some of the leading nonlinear dimensionality reduction algorithms. By creating data generating processes that allow performance to be compared to a reasonable ``Bayes error'' metric, we are able to provide the beginnings of a more rigorous study for dimensionality reduction quality metrics. Based on our observations, we find that metrics based on maximizing the overlap in nearest neighbors such as $Q_{nx}$ and Spectral Overlap tend to have robust performance for 2 dimensional and higher dimensional cases. As shown in the case of LMDS and the Clusters data set, using these methods can provide potentially informative visualizations without tearing the manifold. Finally, we provide an R wrapper for local MDS which is available for download \href{https://github.com/JonathanJohann/lmds}{here}.

For future work, we look to explore a few avenues. First, by defining a unifying metric to go by, this comparison acts as a first step to a fuller study of the existing nonlinear dimensionality reduction algorithms. We also intend to further investigate the realm of dimensionality reduction quality metrics in order to define more intuitive or analytically promising measures of local structure and then global structure. Finally, we aim to develop a method that more directly uses the $Q_{NX}$ and Spectral Overlap quality metric and will potentially become a competitor to leading methods.

\newpage 
\appendix

\section{Results}\label{app:results}

\begin{table}[htbp]
	\scriptsize
	\centering
	\begin{tabular}{|c|c|c|c|c|c|c|c|c|c|}
\hline
Algorithm & Data & $K_{max}$ & $Q_{nx}$ & Entropy & Mutual Info & Local Error & Spectral Error & Spearman & 1-NN\\
\hline
LMDS&Clusters&	0.955&	0.880&	0.688	&0.241&	0.698&	0.911&	0.506&	0.958	\\
tSNE&Clusters&	0.960&	0.968&	0.965&	0.960&	0.965&	0.965&	0.965&	0.975	\\
UMAP&Clusters&	0.964&	0.966&	0.963&	0.959&	0.965&	0.963&	0.783&	0.975	\\
Sammon&Clusters&	0.451&	0.451&	0.451&	0.451&	0.451&	0.451&	0.451&	0.451	\\

\hline 
\end{tabular}
	\caption{High Dimensional Dataset}
	\label{table:clusters.tex}
\end{table}

\begin{table}[htbp]
	\scriptsize
	\centering
	\begin{tabular}{|c|c|}
\hline
Method & Avg. Procrust. Dist.\\
\hline
Local MDS  & 98.26\\
Sammon Mapping & 9.51\\
t-SNE & 103.51\\
UMAP & 78.28\\
\hline
\end{tabular}
	\caption{Procrustes Distances}
	\label{table:algo_summary.tex}
\end{table}
\begin{landscape}
\begin{table}[htbp]
	\scriptsize
	\centering
	\begin{tabular}{|c|c|c|c|c|c|c|c|c|c|}
\hline
Algorithm & Data & $K_{max}$ & $Q_{nx}$ & Entropy & Mutual Info & Local Error & Spectral Error & Spearman & Procrustes\\
\hline
LMDS&	2 Lines&	22.91&	8.73&	8.73&	34.86&	8.73&	8.73&	8.73&	5.42\\
tSNE&	2 Lines&	27.07&	19.79&	19.93&	10.00&	11.57&	19.91&	19.91&	8.28\\
UMAP&	2 Lines&	13.97&	17.06&	13.66&	27.54&	11.24&	17.06&	13.66&	11.14\\
Sammon&	2 Lines&	4.61&	4.61&	4.61&	4.61&	4.61&	4.61&	4.61&	4.61\\
\hline
LMDS&	Circles&	2.05&	2.05&	2.05&	8.93&	2.05&	2.05&	2.05&	0.34\\
tSNE&    Circles&	0.96&	0.51&	0.95&	8.81&	0.94&	0.45&	0.45&	0.45\\	
UMAP&	Circles&	8.91&	1.46&	1.46&	8.58&	4.33&	1.46&	1.46&	1.46\\
Sammon&	Circles&	0.23&	0.23&	0.23&	0.23&	0.23&	0.23&	0.23&	0.23\\
\hline
LMDS&	Trefoil&	5.22&	5.22&	5.22&	24.19&	5.22&	5.22&	5.22&	1.05\\
tSNE&	Trefoil&	1.92&	1.66&	1.66&	24.29&	2.61&	1.55&	1.55&	1.51\\
UMAP&	Trefoil&	9.68&	3.74&	4.37&	26.21&	5.57&	3.74&	3.74&	3.74\\
Sammon&	Trefoil&	0.53&	0.53&	0.53&	0.53&	0.53&	0.53&	0.53&	0.53\\
\hline
LMDS&	Xs&	442.34&	53.39&	53.39&	637.30&	72.36&	63.48&	72.36&	23.09\\
tSNE&	Xs&	900.26&	141.65&	101.41&	146.97&	279.97&	78.61&	78.61&	78.61\\
UMAP&	Xs&	342.33&	211.24&	140.10&	542.88&	91.91&	211.24&	211.24&	60.54\\
Sammon&	Xs&	1.51&	1.51&	1.51&	1.51&	1.51&	1.51&	1.51&	1.51\\
\hline
LMDS&	3 Gaussians&	79.12&	14.55&	38.44&	56.75&	38.59&	38.59&	14.55&	14.55\\
tSNE&	3 Gaussians&	48.56&	48.69&	48.26&	75.47&	48.26&	48.45&	48.45&	46.76\\
UMAP&	3 Gaussians&	47.11&	48.52&	60.76&	78.24&	48.83&	56.97&	47.75&	46.57\\
Sammon&	3 Gaussians&	43.95&	43.95&	43.95&	43.95&	43.95&	43.95&	43.95&	43.95\\

\hline

\end{tabular}
	\caption{Procrustes Distances}
	\label{table:results.tex}
\end{table}
\end{landscape}

%\begin{figure}
%  \includegraphics[width=\linewidth]{figures/tsn%e_Xs_dr_results.png}
%  \caption{t-SNE X's Dataset result}
%  \label{fig:hdclus}
%\end{figure}
%\begin{figure}
%  \includegraphics[width=\linewidth]{figures/uma%p_xs_results.png}
%  \caption{UMAP X's Dataset result}
%  \label{fig:hdclus}
%\end{figure}
%\begin{figure}
%  \includegraphics[width=\linewidth]{figures/lmd%s_Xs.png}
%  \caption{LMDS X's Dataset result}
%  \label{fig:hdclus}
%\end{figure}

\section{Data Generating Process}\label{app:dgp}
Below outlines the mathematical formulation for each of the data generating processes:

\paragraph{2 Lines} The 2 Lines dataset from \cite{wattenberg2016how} is generated as two parallel line segments. This can be done as such:

\begin{alignat*}{2}
    X^{n \times 2} \sim N(0_{2 \times 2},I_{2 \times 2}) &,& m = \lfloor \tfrac{n}{2}  \rfloor\\
    X_{1:m} &=& 2 X_{1:m} + 5\\
    X_{m+1:2m} &=& 2 X_{m+1:2m} - 5\\
\end{alignat*}

\paragraph{3 Gaussians} The 3 Gaussians dataset inspired by \cite{wattenberg2016how} is generated as 3 separable clusters with 2 closer to each other and 1 with large variance. We formulate the data as such:
\begin{alignat*}{2}
    m &=&  \lfloor \tfrac{n}{3}  \rfloor\\
    X_1^{m \times 2} &\sim& N(-5,I_{2 \times 2})\\
    X_2^{m \times 2} &\sim& N(-10,I_{2 \times 2})\\
    X_3^{m \times 2} &\sim& N(20,10 * I_{2 \times 2})
\end{alignat*}
\paragraph{Trefoil Knot} The trefoil knot also comes from \cite{wattenberg2016how} and we add a small amount of noise $\epsilon$. The data generating process is:

\begin{alignat*}{2}
    \phi &=& 0 + (i-1) * \tfrac{\pi}{n}, i = 1,...,n\\
    x &=&sin(\phi) + 2sin(2\phi) + \epsilon\\
    y &=&cos(\phi) - 2cos(2\phi) + \epsilon\\
    \epsilon &\sim& N(0,0.01 * I_{2 \times 2})\\
    X &=& \begin{bmatrix}
    \phi & x\\
    \phi & y
    \end{bmatrix}\\
\end{alignat*}
where our final input matrix is $X$.

\paragraph{Curved X's} The Curved X's data set is inspired by \cite{hastie2009elements} who show how using linear methods does not often guarantee that the nearest neighbor relationships are kept and that nonlinear methods can do a better job by focusing on local structure. This data set is a slight change and consists of two curved X's which are specified as such:
\begin{align*}
    X_1 = [-2.1147,2.1147]\qquad 
    X_2 = [-2.1147,2.1147] \\
    Y_1 = 50 + sin(10\pi x) \cdot x^4\qquad
    Y_2 = -50 - sin(10\pi x) \cdot x^4\\
    X = \begin{bmatrix}
    X_1\\
    X_2
    \end{bmatrix}\qquad
    Y = \begin{bmatrix}
    Y_1\\
    Y_2
    \end{bmatrix}\\
    C = [X,Y] \cdot \begin{bmatrix}
    cos(\pi/4) & sin(\pi/4)\\
    -sin(\pi/4) & cos(\pi/4)
    \end{bmatrix}
\end{align*}
where our final input matrix is $C$.

\paragraph{Noisy Circles} The Noisy Circles data set comes from \cite{sklearn2018comparing} and is one circle drawn inside a larger circle. We also add noise and the data generating process is therefore as such:
\begin{alignat*}{2}
    \phi &=& 0 + (i-1) * \tfrac{4\pi}{n}, i = 1,...,n\\
    x &=&\ [\ 0.5*cos(\phi)+\epsilon,cos(\phi)+\epsilon\ ] \\
    y &=&\ [\ 0.5*sin(\phi)+\epsilon,sin(\phi)+\epsilon\ ] \\
    \epsilon &\sim& N(0,0.0004 * I_{2 \times 2})
\end{alignat*}

\paragraph{High Dimensional Clusters} For the high dimensional clusters data set, we place clusters on the corners of a 4-dimensional hyper-cube as such:
\begin{alignat*}{2}
    \mu &=
    \begin{bmatrix}
    0 & 0 & 0 & 0\\
    0 & 0 & 0 & 5\\
    \vdots  & \vdots & \vdots & \vdots \\
    5 & 5 & 5 & 5\\
    \end{bmatrix}_{16 \times 4}
\end{alignat*}
\begin{alignat*}{2}
    x_1 & \sim & N(\mu_{1,\cdot},I_{4 \times 4})\\
    & \vdots & \\
    x_{16} & \sim & N(\mu_{16,\cdot},I_{4 \times 4})\\
    X & = & [x_1,...,x_{16}]\\
\end{alignat*}

\section{Quality Metrics}\label{app:metrics}

Below we outline the formulation for each of the quality metrics:

\paragraph{$K_{max}$ and $Q_{nx}$ Quality Metrics}

$K_{max}$ was proposed in \cite{lee2009quality} and chooses the maximum LCMC score from \cite{chen2009local} for all available K. The metric begins by calculating $Q_{NX}$ from \cite{liang2017new} and is calculated as:
\begin{equation}
Q_{NX}(K) = \tfrac{1}{KN} \sum_{k=1}^K \sum_{l=1}^K Q_{kl}
\end{equation}
which counts the number of points that remained in the same local neighborhood defined by $k$. This can also be conceptualized as penalizing the number of mismatched ranks when comparing ranked Euclidean distances in high and low dimensional space. One can tweak the number of nearest neighbors $K$ or $K = N-1$ as mentioned in \cite{liang2017new}, removing the tuning element. Next, the LCMC is calculated as:
\begin{equation}
LCMC(K) = Q_{NX}(K) - \frac{K}{N-1}
\end{equation}
And finally, we achieve our choice of $K_{max}$ via:

\begin{equation}
K_{max} = argmax_{K} LCMC(K) = argmax_{K} \big( Q_{NX}(K) - \frac{K}{N-1} \big)
\end{equation}

\paragraph{Entropy and Mutual Information}

Both methods come from \cite{babaee2013assessment} and first require the user to calculate the co-ranking matrix as follows:
\begin{equation}
M = [m_{kl}]_{1\leq k,l\leq N-1}
\end{equation}
for all N data points and 
\begin{equation}
m_{kl} = | \{(i,j) : (A_{ij} = k) \cap (B_{ij} = l)\}|
\end{equation}
Next, the joint probability distribution $P(i,j)$ is specified as follows:
\begin{equation}
P(i,j) = \tfrac{1}{N-1} M
\end{equation}
The \textbf{Entropy} quality metric is: 
\begin{equation}
H = - \sum_i \sum_j P(i,j) log P(i,j)
\end{equation}
The \textbf{Mutual Information} metric is:
\begin{equation}
MI = \sum_i \sum_j P(i,j) log \frac{P(i,j)}{P(i) P(j)}
\end{equation}

\paragraph{Local Error}

Similar to Kruskal Stress, we devise a meta criterion that focuses more on getting the correct immediate errors by linearly weighting lower ranked neighbors higher than 
higher ranked neighbors. The result is a sum over the cumulative sum of the squared distance errors. More succinctly, we can think of this as linearly weighting the squared distance errors as:
\begin{align}
\text{Local Error} = \sum_{i}^{N-1} \sum_{j}^{N} \sum_{k\in N(i)} ||D_{jk} - d_{jk}||^2_2
\end{align}
where $D_{jk}$ and $d_{jk}$ are the Euclidean distance in high and low dimensional space and $N(i)$ is the neighborhood for a given point defined by the $i$th nearest neighbors.

\paragraph{Spectral Overlap} Let $\tilde{K}^i$ be the KNN graph in input space $X$ with $k$ parameter $i$ and $K^i$ be the KNN graph in output space $Y$ with $k$ parameter $i$. The metric is calculated as:
\begin{align}
    \text{Spectral Overlap} = 1 - \sum_{i=1}^{n-1} \tfrac{1}{i\cdot (n-1)} \sum_{j=1}^{(n-1)} \sum_{k\not = j} \tilde{K}^i_{jk} * K^i_{jk}
\end{align}
The intuition here is that we want to have overlap in every KNN graph for $k=1,...,n-1$. This provides more weight on more immediate neighbors and results in a light penalty if pairwise neighbor relations are off by one or two ranks. However, if there is a drastic tear in the manifold, this will more heavily penalize terms that should be a nearest neighbor to a given point but are very far from said given point.
%{\color{red} This is the same as ranking each column of the input and output matrix in \textit{descending} order, dividing by the L1 norm of the ranked matrix, and then subtracting this value from 1.}

\paragraph{Spearman} The Spearman rank correlation coefficient is calculated on all values of $X$ and $Y$ for one coefficient.  

\paragraph{Procrustes Distance} From \cite{hastie2009elements}, the objective is the l2 norm on the rotated, shifted, and scaled image $Y$ with respect to the reference image $X$.
\begin{align}
    \min_{R,\beta} ||X - \beta \cdot Y \cdot R||_F
\end{align}

\paragraph{1-NN} From \cite{van2009dimensionality}, one metric that can be used when the true groups are known is the accuracy of the first nearest neighbor classifier. We can calculate this quickly via the following process. Let $B$ be the column-wise rank of each element $Y_{ij}$ excluding elements where $i=j$ which have default distance $0$. Let each class $i$ of $k$ total classes have $n_i$ elements in $X$.

\begin{align}
    G = \mathbf{1}\{B_{ij} = 1\}\qquad  \forall B_{ij} \in B\\
    H = \textbf{blkdiag}(\mathbf{1}^{n_1\times n_1},...,\mathbf{1}^{n_k\times n_k})
\end{align}
where \textbf{blkdiag} corresponds to a block diagonal matrix of diagonal blocks of size $n_1,...,n_k$ such that $n_1+...+n_k = n$ or the total number of points in the input matrix $X$.

We then calculate $1-NN$ as such:

\begin{align}
    \text{1-NN} = \tfrac{1}{n} \sum_{i=1}^{n} \sum_{j=1}^n G_{ij} \cdot H_{ij}
\end{align}

\bibliographystyle{plainnat}
\bibliography{main}

\end{document}